\documentclass[runningheads]{llncs}
\usepackage[T1]{fontenc}
\usepackage[utf8]{inputenc}
\usepackage{textgreek}
\usepackage{amsmath}
\usepackage{booktabs}
\usepackage{graphicx}
\usepackage{multirow}
\usepackage{makecell}
\usepackage{threeparttable}
\usepackage[table]{xcolor}   
\usepackage{amssymb}          
\usepackage{graphicx,verbatim}
\definecolor{ours}{gray}{0.9}   
\DeclareUnicodeCharacter{03BA}{\textkappa}
\usepackage[sort&compress,numbers]{natbib}

\makeatletter
\renewcommand\@biblabel[1]{#1.} 
\makeatother
\begin{document}
\title{PPGL-Swarm: Integrated Multimodal Risk Stratification and Hereditary Syndrome Detection in Pheochromocytoma and Paraganglioma}
\titlerunning{PPGL-Swarm}
%

\author{
Zelin Liu\inst{1}\textsuperscript{*} \and
Xiangfu Yu\inst{2}\textsuperscript{*} \and
Jie Huang\inst{3} \and
Ge Wang\inst{1} \and
Yizhe Yuan\inst{1} \and
Zhenyu Yi\inst{1} \and
Jing Xie\inst{3} \and
Haotian Jiang\inst{4} \and
Lichi Zhang\inst{1}\textsuperscript{$\dagger$}
}

\authorrunning{Z. Liu and X. Yu et al.}

\institute{
Shanghai Jiao Tong University \and
The Chinese University of Hong Kong \and
Ruijin Hospital \and
ShanghaiTech University
}

\maketitle

\begingroup
\renewcommand{\thefootnote}{}
\footnotetext{\hspace{-1.5em}\textsuperscript{*} These authors contributed equally.}
\footnotetext{\hspace{-1.5em}\textsuperscript{$\dagger$} Corresponding author: Lichi Zhang. Email: lichizhang@sjtu.edu.cn}
\endgroup
\begin{abstract}
Pheochromocytomas and paragangliomas (PPGLs) are rare neuroendocrine tumors,of which 15–25\% develop metastatic disease with 5--year survival rates reported as low as 34\%. PPGL may indicate hereditary syndromes requiring stricter, syndrome-specific treatment and surveillance, but clinicians often fail to recognize these associations. Clinical practice uses GAPP score for PPGL grading, but several limitations remain: (1) GAPP scoring demands a high workload for clinician because it requires the manual evaluation of six independent components; (2) key components such as cellularity and Ki-67 are often evaluated with subjective criteria; (3) Several clinically relevant metastatic risk factors are not captured by GAPP, such as SDHB mutations, which have been associated with reported metastatic rates of 35--75\%. Agent-driven diagnostic systems appear promising, but most lack traceable reasoning for decision-making and do not incorporate domain-specific knowledge such as PPGL genotype information.To address these limitations, we present PPGL-Swarm, an agentic PPGL diagnostic system that generates a comprehensive report, including automated GAPP scoring (with quantified cellularity and Ki-67), genotype risk alerts, and multimodal report with integrated evidence. The system provides an auditable reasoning trail by decomposing diagnosis into micro-tasks, each assigned to a specialized agent. The gene and table agents use knowledge enhancement to better interpret genotype and laboratory findings, and during training we use reinforcement learning to refine tool selection and task assignment. Our model achieves state-of-the-art performance in report quality, GAPP accuracy, and mutation prediction accuracy.


\end{abstract}
\section{Introduction}
\noindent 

\begin{figure}[htb]
\centering
\includegraphics[width=0.8\textwidth]{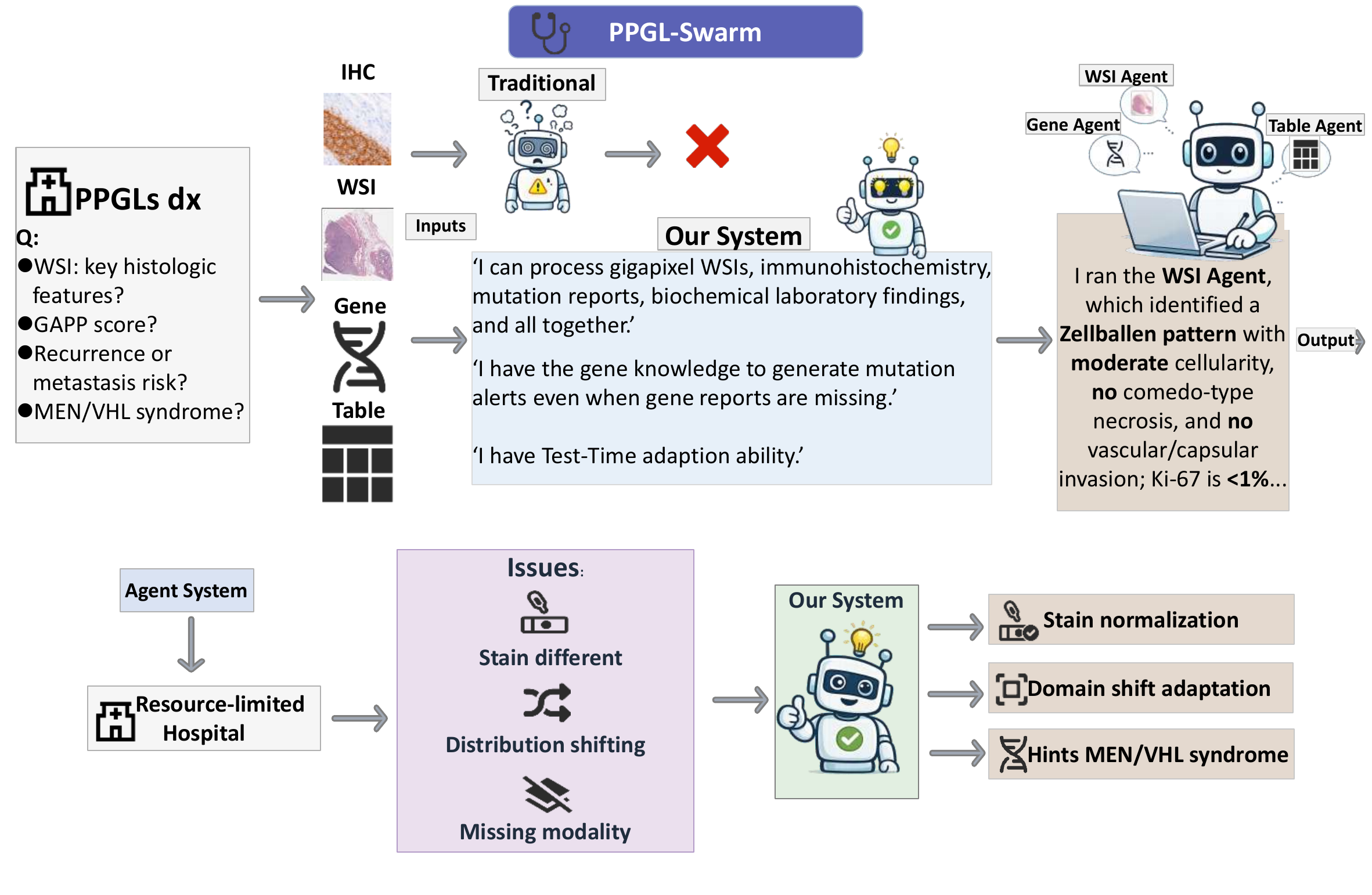}
\caption{Our system can orchestrate agents to process gigapixel WSI/IHC and mutimodal data.}
\label{fig1}
\end{figure}

Pheochromocytomas and paragangliomas (PPGLs) are neuroendocrine tumors with an estimated incidence of 6 cases per 1,000,000 person-years. Despite this rarity, up to 15-25\% of PPGLs develop into metastatic disease, which carries a dismal prognosis with reported 5-year survival rates as low as 34\%\cite{WachtelHeather2020PMPi_v2,IlanchezhianMaran2020ETfA}. Recent WHO guidelines have shifted PPGL classification from binary benign/malignant designations to multidimensional risk assessment, recognizing that all PPGLs carry metastatic potential\cite{MeteOzgur2022Oot2,KohJung-Min2017Vopg}. The GAPP scoring system operationalizes this approach by integrating histologic pattern, cellularity, comedo-type necrosis, capsular/vascular invasion, catecholamine type, and Ki-67 index to predict metastatic risk\cite{KimuraNoriko2014Pgfp}. However, clinical adoption of GAPP is hindered by two practical limitations: its reliance on manual region selection, and quantitative cell counting (requiring up to 500 cells per case) is labor-intensive, while subjective interpretation of several parameters introduces inter-observer variability and limits predictive reproducibility\cite{LiQin2024VaEo,WachtelHeather2020PMPi_v2}.


However, GAPP itself has inherent limitations because it does not incorporate several established prognostic factors. Genotype is paramount, as SDHB mutations confer a metastatic risk of 35–75\% and are associated with markedly shorter median survival (42 vs 244 months)\cite{EijkelenkampKarin2020Ciot,HescotSegolene2019PoMP,NöltingSvenja2022PMoP}. In addition, 33.8\% of PPGL patients carry germline mutations, including VHL (7.3\%) and RET (6.3\%), which define hereditary syndromes requiring distinct clinical management.\cite{Lenders2014JCEM} Moreover, larger tumor size is positively correlated with metastatic risk, and bilateral disease is more frequent in hereditary cases and therefore suggests a syndromic etiology. In clinical practice, these genotype-phenotype associations may remain unrecognized despite affecting up to one-third of patients, owing to limited awareness or restricted access to genetic testing.

Agentic LLM systems have emerged as a promising paradigm for 
complex clinical reasoning, as they can decompose multistep 
diagnostic workflows, coordinate specialized tools\cite{Gorenshtein2025.08.22.25334232, ChenXi2025Edcw, GoodellAlexJ.2025Llma, SilveiraAndreLehdermann2026Msfc}. However, applying such systems 
to PPGL diagnosis exposes several limitations. First, existing medical agents are unable to jointly reason over whole slide images, immunohistochemistry, mutation reports, and biochemical laboratory findings within a unified diagnostic framework\cite{LuMingY.2024AmgA, AlSaadRawan2024MLLM}. Second, these systems lack PPGL knowledge: they cannot interpret the prognostic implications of mutations such as SDHB, nor associate biochemical phenotypes with syndromic genotypes such as VHL or RET, making them ill-suited for hereditary risk estimation in routine care\cite{TaïebDavid2024Mopa}. 
Third, 
Existing WSI diagnostic models, such as SlideChat\cite{chen2024slidechat} and GIANT\cite{buckley2025navigatinggigapixelpathologyimages}, focus on processing high-resolution pathology images\cite{ding2025multimodal}. In contrast, we decompose the overall diagnostic task into fine-grained, clinically defined subtasks, and subsequently integrate their outputs using a dedicated reasoning model. This structured paradigm is more reliable and facilitates the incorporation of advanced computational methods.\cite{KimuraNoriko2014Pgfp}. 
Fourth, deployment across 
resource-limited institutions is further undermined by staining 
variability and distribution shift\cite{MeteOzgur2022Oot2}. 

Together, these gaps call for a system capable of comprehensive multimodal reasoning across pathology, genetics, and laboratory data, equipped with structured domain knowledge for mutation interpretation, and able to provide fine-grained, verifiable diagnostic outputs with 
genotype-aware risk alerts that can support clinicians who may 
be unaware of hereditary associations or lack access to genetic 
testing.

To address these limitations, we present PPGL-Swarm, an agentic system for comprehensive PPGL diagnosis(Fig.~\ref{fig1}). Our contributions 
are three-fold: (1) To enable unified multimodal reasoning across 
heterogeneous clinical inputs, we propose an agent swarm 
architecture in which a central decision agent orchestrates dedicated 
WSI, gene, and table swarms over whole slide images, 
immunohistochemistry, mutation reports, and biochemical laboratory 
findings, with reinforcement learning applied to optimize tool 
selection and swarm coordination; (2) To support reliable 
genotype-aware hereditary risk estimation without hallucinated 
associations, we introduce a structured knowledge graph retrieval 
mechanism for the gene and table swarms, grounding the interpretation 
of genetic variants and biochemical phenotypes to generate actionable 
risk alerts for VHL and MEN2 syndromes; (3) To support diagnostic 
transparency, we decompose diagnosis into micro-tasks aligned with individual GAPP criteria, 
producing independently verifiable outputs for each pathological 
component including quantified Ki-67 and cellularity, complemented 
by a test-time adaptation strategy that accommodates 
heterogeneous staining protocols across resource-limited institutions.
\section{Methodology}

\begin{figure}[htb]
\centering
\includegraphics[width=0.8\textwidth]{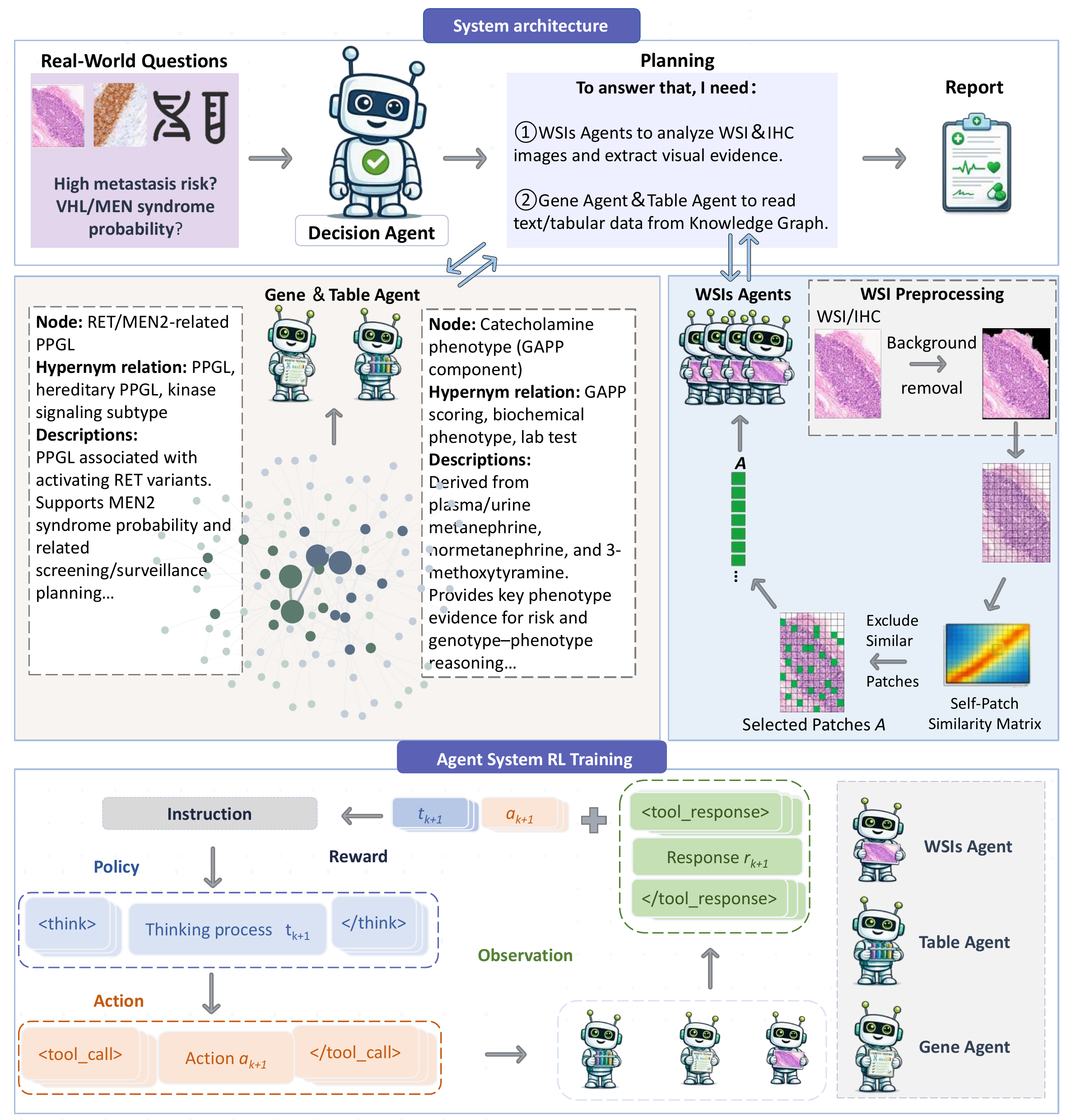}
\caption{Our architecture and the RL training} 
\label{fig2}
\end{figure}

The system comprises a central decision agent (Qwen3.5-35B-A3B) and three specialized swarms: a WSI swarm for histological image analysis, a Table swarm for biochemical laboratory data, and a Gene swarm for mutation interpretation (Fig.~\ref{fig2}). The central agent receives the diagnostic instruction, dispatches sub-tasks to each swarm and synthesizes the returned evidence into a unified report. It is trained via reinforcement learning to optimize swarm coordination.

\noindent\textbf{WSI Swarm.}
The WSI swarm evaluates five targets aligned with the visual components of GAPP scoring: histologic pattern (zellballen, large irregular nests, or pseudorosette), comedo-type necrosis, vascular/capsular invasion, cellularity, and Ki-67 index. The first three are classification tasks. Cellularity and Ki-67 are formulated as regression tasks that output continuous quantitative values rather than the coarse ordinal grades used in clinical practice, addressing the subjectivity.

Patch features are extracted by a frozen UNI-v2 encoder $f_\phi$ and aggregated by a fully trainable TransMIL aggregator $g_\psi$, producing slide representation $h = g_\psi(\{f_\phi(p_i)\}_{i=1}^{N})$\cite{chen2024uni, NEURIPS2021_10c272d0}. The aggregator is first trained on histologic pattern classification, after which its weights are frozen. Each remaining target is fine-tuned by training only a task-specific head $k_j$ on top of the fixed $h$, using cross-entropy loss for classification targets and mean squared error for regression targets. This protocol preserves the shared representation while preventing overfitting on targets with limited supervision.

To handle staining variability, inference applies an adaptation. 
First, each test slide is colour-normalised in CIE LAB space by aligning its tissue-pixel statistics to the training distribution:
\begin{equation}
    I_\text{out} = \sigma_\text{tgt} \cdot \frac{I_\text{src} - \mu_\text{src}}{\sigma_\text{src} + \epsilon} + \mu_\text{tgt}
\end{equation}
where $(\mu_\text{src}, \sigma_\text{src})$ and $(\mu_\text{tgt}, \sigma_\text{tgt})$ are the per-channel mean and standard deviation of the test slide and training set respectively, computed over tissue pixels in LAB space. 

Second, during the subsequent forward pass, the running statistics of all batch normalisation layers are updated via AdaBN:
\begin{equation}
    \hat{\mu} = (1-\alpha)\,\mu_\text{train} + \alpha\,\mu_\text{current}, \qquad
    \hat{\sigma}^2 = (1-\alpha)\,\sigma^2_\text{train} + \alpha\,\sigma^2_\text{current}
\end{equation}
where $\mu_\text{current}$ and $\sigma_\text{current}$ are computed from the current test slide activations and $\alpha$ is the momentum parameter. 
Although inference uses a batch size of 1, statistics remain stable as they are estimated over high-dimensional spatial feature activations, and the running estimates are progressively accumulated across test samples. 

Normalisation removes inter-site chromatic variation; AdaBN then corrects residual feature-level distribution shift, together enabling robust cross-institutional deployment without any labelled target-domain data.

\noindent\textbf{Gene Swarm and Table Swarm.}
Both swarms augment Qwen3.5-35B-A3B reasoning via a structured knowledge graph whose nodes encode PPGL-relevant entities, including genetic variants, biochemical phenotypes, and syndromic classifications, each formalised by an entity name, a hypernym relation to broader biological categories, and a description covering pathway associations and clinical implications. When a mutation or biochemical indicator is queried, the swarm retrieves the corresponding node and passes it to the central agent, grounding interpretation and preventing hallucinated associations.

The Gene swarm additionally incorporates three binary mutation prediction heads (SDHB, VHL, RET) built on the same frozen $f_\phi$ and $g_\psi$ from the WSI swarm, following the identical fine-tuning protocol. Each head outputs a confidence score $c_m \in [0,1]$ for mutation $m$, enabling genotype-aware risk alerts from histological appearance alone when molecular testing results are unavailable.

The Table swarm processes biochemical laboratory findings, including catecholamine phenotype derived from plasma metanephrine, normetanephrine, and 3-methoxytyramine, which corresponds to the sixth GAPP component not captured by WSI.

\noindent\textbf{Reinforcement Learning for Central Agent.}
\label{sec:rl}
The central decision agent is modelled as a policy $\pi_\theta(a_t \mid s_t)$, where $s_t$ encodes the diagnostic instruction and the history of observations returned by swarms up to step $t$, and $a_t$ is a tool call to one of the three swarms. The agent first generates an internal reasoning trace, executes $a_t$, receives the swarm observation, and iterates until it produces the final report. This interaction is cast as a Markov decision process, and the objective is to maximise the expected discounted return:
\begin{equation}
    J(\theta) = \mathbb{E}_{\tau \sim \pi_\theta}\!\left[\sum_{t=0}^{T} \gamma^t R(s_t, a_t)\right]
\end{equation}
where $\gamma \in [0,1]$ is the discount factor and $\tau = (s_0, a_0, s_1, a_1, \ldots)$ is a sampled trajectory. The reward decomposes as:
\begin{equation}
    R(s_t, a_t) = r_\text{diag}(s_t, a_t) + \lambda_1 \cdot \mathbb{1}_\text{format}(a_t) - \lambda_2 \cdot \mathbb{1}_\text{redundant}(a_t)
\end{equation}
where $r_\text{diag}$ measures diagnostic correctness against ground-truth GAPP scores and genotype labels, $\mathbb{1}_\text{format}$ penalises malformed tool calls, and $\mathbb{1}_\text{redundant}$ penalises unnecessary swarm invocations. $\lambda_1$ and $\lambda_2$ are scalar weighting coefficients. The policy is optimised via the policy gradient:
\begin{equation}
    \nabla_\theta J(\theta) \approx \frac{1}{N}\sum_{i=1}^{N}\sum_{t=0}^{T} \nabla_\theta \log \pi_\theta(a_t^{(i)} \mid s_t^{(i)})\,\hat{A}_t^{(i)}
\end{equation}
where $N$ is the batch size and $\hat{A}_t^{(i)}$ is the advantage estimate at step $t$ for trajectory $i$, computed via generalised advantage estimation (GAE). By maximising $J(\theta)$, the agent learns to invoke swarms in a diagnostically efficient order and to synthesize their outputs into accurate, well-structured reports.

\begin{figure}[htb]
\centering
\includegraphics[width=\textwidth]{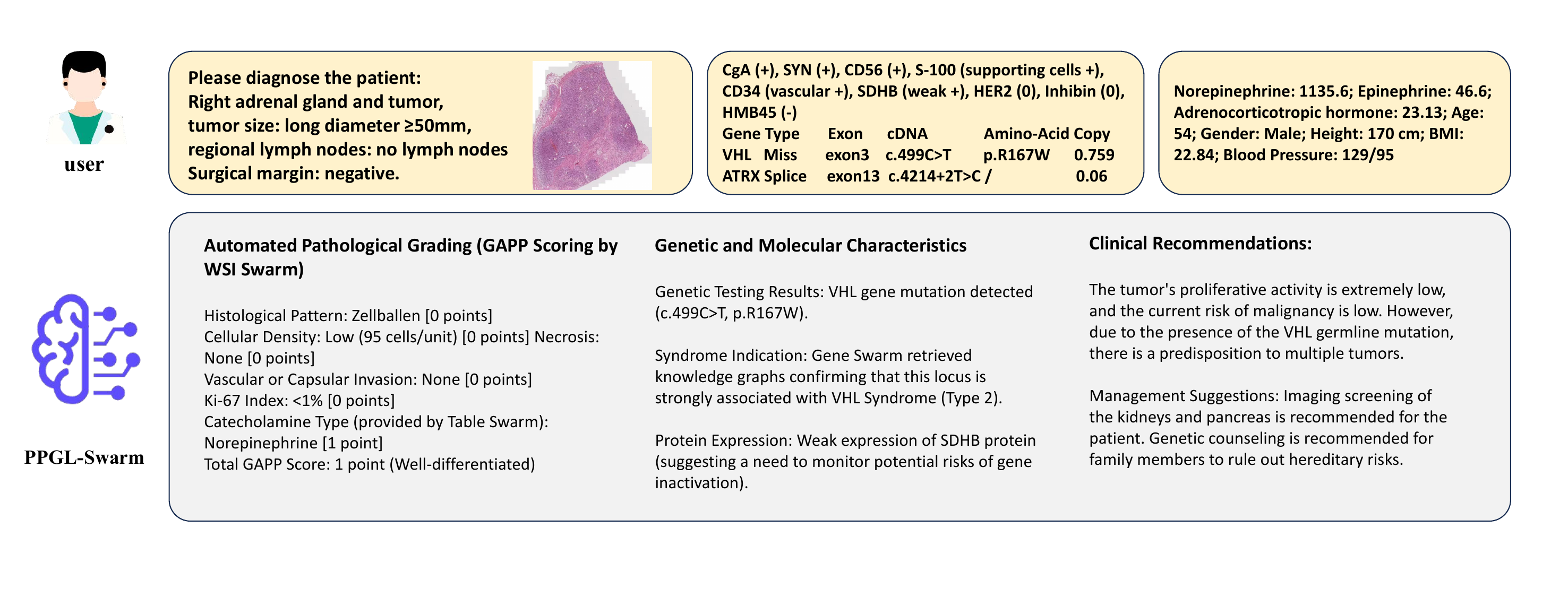}
\caption{Visualized results of our system’s performance.}
\label{fig3}
\end{figure}

\section{Experiments}

\begin{table*}[ht]
\centering
\caption{Overall diagnostic report quality.}
\label{tab:report}
\setlength{\tabcolsep}{5pt}

\resizebox{\textwidth}{!}{
\begin{tabular}{llccccc}
\toprule
\multirow{2}{*}{Model} & \multirow{2}{*}{Input} &
\multicolumn{4}{c}{Report Quality (1--4\,{\small$\uparrow$})} &
\multirow{2}{*}{\makecell{Overall\\{\small$\uparrow$}}} \\
\cmidrule(lr){3-6}
& & \makecell{Diagnostic\\Accuracy} & \makecell{Clinical\\Actionability} & Completeness & Format & \\
\midrule


GPT-4o & Patch+Gene+Table 
& $2.8{\pm}0.4$ & $2.3{\pm}0.5$ & $2.1{\pm}0.3$ & $3.1{\pm}0.4$ 
& 2.58 \\

GPT-4o & Thumbnail+Gene+Table 
& $2.2{\pm}0.3$ & $2.0{\pm}0.4$ & $1.9{\pm}0.3$ & $3.0{\pm}0.5$ 
& 2.28 \\

Claude-4.5-Sonnet & Patch+Gene+Table  
& $2.7{\pm}0.3$ & $2.2{\pm}0.4$ & $2.0{\pm}0.4$ & $3.2{\pm}0.3$ 
& 2.53 \\

\midrule

LLaVA-Med & Patch+Gene+Table  
& $2.1{\pm}0.5$ & $1.8{\pm}0.4$ & $1.7{\pm}0.3$ & $2.6{\pm}0.4$ 
& 2.05 \\

MedDr & Patch+Gene+Table  
& $2.9{\pm}0.4$ & $2.5{\pm}0.3$ & $2.3{\pm}0.4$ & $3.2{\pm}0.3$ 
& 2.73 \\

SlideChat & Slide+Gene+Table 
& $3.1{\pm}0.4$ & $\underline{2.8{\pm}0.4}$ & $\underline{2.6{\pm}0.3}$ & $\underline{3.3{\pm}0.3}$ 
& \underline{2.95} \\

TITAN & Slide+Gene+Table  
& $\underline{3.4{\pm}0.3}$ & $2.6{\pm}0.4$ & $2.4{\pm}0.3$ & $3.1{\pm}0.4$ 
& 2.88 \\

\midrule
\rowcolor{ours}
\textbf{Ours} & Slide+Gene+Table 
& $\mathbf{3.6{\pm}0.3}$ & $\mathbf{3.0{\pm}0.3}$ & $\mathbf{2.7{\pm}0.2}$ & $\mathbf{3.5{\pm}0.3}$ 
& \textbf{3.2} \\

\bottomrule
\end{tabular}
}

\end{table*}

\noindent\textbf{Dataset.}
We collected 268 patients and 1,168 whole slide images (WSIs). Each case includes immunohistochemistry (IHC) reports, germline mutation profiles, biochemical findings. Ground-truth GAPP scores were determined by two board-certified pathologists. Discrepancies were resolved by consensus to produce a single final score per case. Diagnostic reports were dictated by senior pathologists and verified by a second reviewer, forming the reference standard. Comprehensive diagnostic reports were compiled by senior attending pathologists.

\noindent\textbf{Implementation Details.}
All experiments were conducted on 16 NVIDIA H20 GPUs. Patches of size $256 \times 256$ were extracted at $20\times$ magnification. The TransMIL aggregator was trained for histologic pattern classification for 30 epochs using Adam (learning rate $1 \times 10^{-4}$, weight decay $1 \times 10^{-5}$). Classification tasks used cross-entropy loss; regression tasks used mean squared error. For test-time adaptation, AdaBN was applied with momentum $\alpha = 0.1$. The central decision agent was initialised from Qwen3.5-35B-A3B and optimised using policy gradient with $\gamma = 0.95$, $\lambda = 0.97$, $\lambda_1 = 0.1$, and $\lambda_2 = 0.2$. Training used batch size 16 trajectories. Results are averaged over five-fold cross-validation.

\begin{table}[ht]
\centering
\caption{
Multi-task performance on GAPP scoring and gene mutation prediction.}
\label{tab:gapp}
\setlength{\tabcolsep}{4.5pt}

\resizebox{\textwidth}{!}{
\begin{tabular}{llccc cc cc cc}
\toprule
\multirow{2}{*}{Model} & \multirow{2}{*}{Input} &
\multicolumn{3}{c}{Classification (macro-F1\,{\small$\uparrow$})} &
\multicolumn{2}{c}{Cellularity} &
\multicolumn{2}{c}{Ki-67 Index} &
\multirow{2}{*}{\makecell{GAPP Total\\MAE\,{\small$\downarrow$}}} &
\multirow{2}{*}{\makecell{Gene Mutation\\F1\,{\small$\uparrow$}}} \\
\cmidrule(lr){3-5}\cmidrule(lr){6-7}\cmidrule(lr){8-9}
& & \makecell{Hist.\\Pattern} & Necrosis & \makecell{Vasc./Cap.\\Invasion} &
MAE\,{\small$\downarrow$} & $r$\,{\small$\uparrow$} &
MAE\,{\small$\downarrow$} & $r$\,{\small$\uparrow$} & & \\
\midrule


GPT-4o            & Patch     & 43.3 & 47.2 & 40.6 & 18.2 & 0.31 & 2.41 & 0.28 & 2.8 & 35.2 \\
GPT-4o            & Thumbnail & 21.6 & 12.2 &  7.0 & 22.5 & 0.18 & 3.10 & 0.15 & 3.5 & 22.1 \\
Claude-4.5-Sonnet & Patch     & 30.3 & 55.3 & 49.2 & 17.8 & 0.33 & 2.28 & 0.32 & 2.6 & 38.4 \\

\midrule

LLaVA-Med         & Patch     & 27.6 & 30.1 & 26.3 & 20.1 & 0.22 & 2.89 & 0.20 & 3.2 & 28.7 \\
MedDr             & Patch     & 57.8 & 33.7 & 54.4 & 14.3 & 0.48 & 1.92 & 0.45 & 2.1 & 51.3 \\
SlideChat         & Slide     & \underline{73.3} & 54.1 & 60.2 & 12.8 & 0.55 & 1.74 & 0.52 & 1.8 & 58.6 \\
TITAN             & Slide     & 71.5 & \underline{60.4} & \underline{65.8} & \underline{9.2} & \underline{0.71} & \underline{1.31} & \underline{0.68} & \underline{1.4} & \underline{65.3} \\

\midrule
\rowcolor{ours}
\textbf{Ours}     & Slide     & \textbf{73.4} & \textbf{61.6} & \textbf{67.3} & \textbf{8.4} & \textbf{0.73} & \textbf{1.27} & \textbf{0.71} & \textbf{1.2} & \textbf{67.8} \\

\bottomrule
\end{tabular}
}

\end{table}

\noindent\textbf{Comparison with State-of-the-Art Methods.}
Following SlideChat~\cite{chen2024slidechat}, we evaluate models without native WSI support using sampled patches or thumbnails, and assess our system on report quality and multi-task prediction.

\textit{Diagnostic Report Quality.}
Three board-certified pathologists rated generated reports on a 4-point Likert scale across four dimensions: Diagnostic Accuracy, Clinical Actionability, Completeness, and Format. As shown in Table~\ref{tab:report}, our method achieves the highest scores in all dimensions, with an overall mean of 3.20, outperforming the strongest baseline (SlideChat, 2.95). Gains are largest in Clinical Actionability (+0.2) and Diagnostic Accuracy (+0.2 over TITAN), indicating improved genotype-aware reasoning and coordinated decision-making.

\begin{table}[htb]
\centering
\caption{Ablation study on diagnostic report quality.}
\label{tab:ablation_report}

\resizebox{0.8\textwidth}{!}{
\begin{tabular}{lccccc}
\toprule
Variant
  & \makecell{Diagnostic \\ Accuracy $\uparrow$}
  & \makecell{Clinical \\ Actionability $\uparrow$}
  & Completeness $\uparrow$
  & Format $\uparrow$
  & Overall $\uparrow$ \\
\midrule

Full Model
  & \textbf{3.6$\pm$0.3} & \textbf{3.0$\pm$0.3} & \textbf{2.7$\pm$0.2} & \textbf{3.5$\pm$0.3} & \textbf{3.20} \\

w/o Knowledge Graph
  & 3.2$\pm$0.4 & 2.3$\pm$0.4 & 2.4$\pm$0.3 & 3.4$\pm$0.3 & 2.83 \\

w/o RL
  & 3.3$\pm$0.4 & 2.6$\pm$0.3 & 2.4$\pm$0.3 & 3.1$\pm$0.4 & 2.85 \\

w/o Gene \& Table Swarm
  & 3.1$\pm$0.4 & 2.1$\pm$0.4 & 2.2$\pm$0.3 & 3.3$\pm$0.3 & 2.68 \\

\bottomrule
\end{tabular}
}

\end{table}

\textit{Multi-task GAPP Scoring and Mutation Prediction.}
GAPP component scores assigned by pathologists serve as ground truth. Classification tasks (Histologic Pattern, Necrosis, Vascular/Capsular Invasion) are evaluated using macro-F1. Regression tasks (Cellularity, Ki-67) use mean absolute error (MAE) and Pearson correlation $r$. Gene mutation prediction (SDHB, VHL, RET) is evaluated by macro-F1. As shown in Table~\ref{tab:gapp}, our method achieves the best performance across all metrics, including GAPP total MAE of 1.2 and gene mutation F1 of 67.8\%, compared to 1.4 and 65.3\% for TITAN. Improvements are consistent across classification and regression tasks.

\begin{table}[htb]
\centering
\caption{Ablation study on multi-task performance.}
\label{tab:ablation_multitask}
\resizebox{\textwidth}{!}{%
\begin{tabular}{lccccccccc}
\toprule
\multirow{2}{*}{Variant}
  & \multicolumn{3}{c}{Classification (macro-F1 $\uparrow$)}
  & \multicolumn{2}{c}{Cellularity}
  & \multicolumn{2}{c}{Ki-67 Index}
  & \multirow{2}{*}{\makecell{GAPP Total \\ MAE $\downarrow$}}
  & \multirow{2}{*}{\makecell{Gene Mut. \\ F1 $\uparrow$}} \\
\cmidrule(lr){2-4}\cmidrule(lr){5-6}\cmidrule(lr){7-8}
  & Hist.\ Pattern & Necrosis & Vasc./Cap.\ Inv.
  & MAE $\downarrow$ & $r$ $\uparrow$
  & MAE $\downarrow$ & $r$ $\uparrow$ & & \\
\midrule
Full Model
  & \textbf{73.4} & \textbf{61.6} & \textbf{67.3}
  & \textbf{8.4}  & \textbf{0.73}
  & \textbf{1.27} & \textbf{0.71}
  & \textbf{1.2}  & \textbf{67.8} \\
w/o AdaBN
  & 71.8 & 59.3 & 65.1
  & 9.1  & 0.68
  & 1.41 & 0.66
  & 1.5  & 67.5 \\
w/o LAB Normalisation
  & 70.3 & 57.8 & 63.4
  & 9.6  & 0.65
  & 1.55 & 0.62
  & 1.7  & 67.4 \\
w/o TTA
  & 67.5 & 54.9 & 60.8
  & 10.4 & 0.60
  & 1.72 & 0.57
  & 2.0  & 67.1 \\
Single Agent
  & 65.8 & 53.2 & 59.7
  & 11.0 & 0.57
  & 1.83 & 0.54
  & 2.3  & 60.2 \\
\bottomrule
\end{tabular}%
}
\end{table}

\noindent\textbf{Ablation Study.}
We ablate key components of the system (Tables~\ref{tab:ablation_report},~\ref{tab:ablation_multitask}). Removing the knowledge graph (\textit{w/o Knowledge Graph}) leads to the largest drop in Clinical Actionability (3.0 $\rightarrow$ 2.3). Removing reinforcement learning (\textit{w/o RL}) reduces report quality (3.20 $\rightarrow$ 2.85) and increases GAPP total MAE (1.2 $\rightarrow$ 1.5). Removing Gene \& Table swarms (\textit{w/o Gene \& Table Swarm}) substantially decreases Clinical Actionability (3.0 $\rightarrow$ 2.1) and Completeness (2.7 $\rightarrow$ 2.2). For test-time adaptation, removing both stages (\textit{w/o TTA}) decreases histologic pattern F1 by 5.9 points and increases cellularity MAE from 8.4 to 10.4. LAB normalisation addresses pixel-level chromatic variation, while AdaBN corrects feature-level distribution shift. Replacing the multi-agent architecture with a single agent consistently degrades performance, supporting the benefit of task decomposition.

\section{Conclusion}

Starting from the first principles of real clinical problems, we developed PPGL-Swarm. Users feedback indicates that they value its quantitative Ki-67/density analysis, integrated clinical knowledge, and genotype-based alerts. PPGL is a very rare disease; the cohort of 268 patients with paired multidimensional examination results represents a valuable clinical asset. We hope this asset helps hospitals without genetic testing, supports clinicians unfamiliar with hereditary syndromes, and aids institutions with limited cases of this disease.

%
%
%
\clearpage
\bibliographystyle{splncs04}
\bibliography{ref}
\end{document}